\documentclass[sigconf]{acmart}
\usepackage{times}
\usepackage{latexsym}
\usepackage{amsmath}
\usepackage{subcaption}
\usepackage{multirow}
\usepackage{booktabs}

\usepackage{threeparttable}

\usepackage[table, dvipsnames]{xcolor}	
\usepackage{tcolorbox}
\usepackage{arydshln}

\usepackage{amssymb}
\usepackage{float}
\usepackage{tabularx}
\usepackage{makecell}
\usepackage{colortbl}
\usepackage{array}
\usepackage{algpseudocode}
\usepackage[ruled,linesnumbered,vlined]{algorithm2e}
\usepackage{xcolor}
\usepackage{xspace}
\usepackage{subcaption}
\definecolor{searchblue}{RGB}{0, 176, 240}
\definecolor{answerred}{RGB}{255, 0, 102}
\definecolor{infogold}{RGB}{255, 153, 0}
\AtBeginDocument{%
  }
\newcommand{\ours}{CroSearch-R1} 
\copyrightyear{2026}
\acmYear{2026}
\setcopyright{cc}
\setcctype{by}
\acmConference[SIGIR '26]{Proceedings of the 49th International ACM SIGIR Conference on Research and Development in Information Retrieval}{July 20--24, 2026}{Melbourne, VIC, Australia}
\acmBooktitle{Proceedings of the 49th International ACM SIGIR Conference on Research and Development in Information Retrieval (SIGIR '26), July 20--24, 2026, Melbourne, VIC, Australia}
\acmDOI{10.1145/3805712.3809921}
\acmISBN{979-8-4007-2599-9/2026/07}

\begin{document}

\title{CroSearch-R1: Better Leveraging Cross-lingual Knowledge for Retrieval-Augmented Generation}

\author{Rui Qi}
\orcid{0009-0006-0611-7682}
\affiliation{%
  \institution{School of Computer Science and Technology, Beijing Jiaotong University}
  \city{Beijing}
  \country{China}
}
\email{cherry@bjtu.edu.cn}

\author{Fengran Mo}
\orcid{0000-0002-0838-6994}
\affiliation{%
  \institution{RALI, Université de Montréal}
  \city{Montréal}
  \state{Québec}
  \country{Canada}
}
  \email{fengran.mo@umontreal.ca}

\author{Sijin Lu}
\orcid{0009-0001-8199-5630}
\affiliation{%
 \institution{School of Computer Science and Technology, Beijing Jiaotong University}
  \city{Beijing}
  \country{China}
}
\email{lusijin@bjtu.edu.cn}

\author{Yufeng Chen}
\orcid{0000-0003-0437-6788}
\affiliation{%
  \institution{School of Computer Science and Technology, Beijing Jiaotong University}
  \city{Beijing}
  \country{China}}
  \email{chenyf@bjtu.edu.cn}

\author{Jian-Yun Nie}
\orcid{0000-0003-1556-3335}
\affiliation{%
  \institution{RALI, Université de Montréal}
  \city{Montréal}
  \state{Québec}
  \country{Canada}
}
  \email{nie@iro.umontreal.ca}

\author{Kaiyu Huang}
\authornote{Corresponding author.}
\orcid{0000-0001-6779-1810}
\affiliation{%
  \institution{School of Computer Science and Technology, Beijing Jiaotong University}
  \city{Beijing}
  \country{China}}
  \email{kyhuang@bjtu.edu.cn}

\renewcommand{\shortauthors}{Trovato et al.}

\begin{abstract}

  A multilingual collection may contain useful knowledge in other languages to supplement and correct the facts in the original language for Retrieval-Augmented Generation (RAG). However, the vanilla approach that simply concatenates multiple pieces of knowledge from different languages into the context may fail to improve effectiveness due to the potential disparities across languages.  To better leverage multilingual knowledge, we propose \textbf{\ours}, a search-augmented reinforcement learning framework to integrate multilingual knowledge into the Group Relative Policy Optimization (GRPO) process. In particular, the approach adopts a multi-turn retrieval strategy with cross-lingual knowledge integration to dynamically align the knowledge from other languages as supplementary evidence into a unified representation space. Furthermore, we introduce a multilingual rollout mechanism to optimize reasoning transferability across languages. Experimental results demonstrate that our framework effectively leverages cross-lingual complementarity and improves the effectiveness of RAG with multilingual collections.
\end{abstract}

%\begin{CCSXML}
\begin{CCSXML}
<ccs2012>
   <concept>
       <concept_id>10002951.10003317.10003347.10003348</concept_id>
       <concept_desc>Information systems~Question answering</concept_desc>
       <concept_significance>500</concept_significance>
       </concept>
 </ccs2012>
\end{CCSXML}

\ccsdesc[500]{Information systems~Question answering}


\keywords{Retrieval-Augmented Generation, Multilingualism, Cross-lingual Retrieval}


\maketitle

\section{Introduction}
% \begin{figure}
%     \centering
%     \includegraphics[width=1.0\linewidth]{figure1_multilingual_retrieval_c3recall.png}
%     \caption{Caption}
%     \label{fig:placeholder}
% \end{figure}

% \begin{table}[t]
% \centering
% \small
% \begin{threeparttable}
% \caption{Impact of multilingual knowledge retrieval on cross-lingual performance.}
% \label{tab:cross-lingual_retrieval}
% \setlength{\tabcolsep}{6pt}
% \begin{tabular}{lccc}
% \toprule
% \textbf{Retrieval Setting} & \textbf{Fr} & \textbf{Th} & \textbf{Ar} \\
% \midrule
% Search-R1\\(Monolingual KB)
% & 34.18 / 53.43
% & 16.60 / 29.61
% & 18.55 / 40.59 \\
% (Local + English KB)
% & 44.82 / 62.25
% & 23.24 / 35.51
% & 18.65 / 42.18 \\
% \bottomrule
% \end{tabular}
% \begin{tablenotes}\footnotesize
% \item \textbf{Note.} Each cell reports \textit{Fem / C3Recall}.
% \end{tablenotes}
% \end{threeparttable}
% \end{table}

% Retrieval-augmented generation (RAG) has emerged as a powerful solution to incorporate external knowledge into Large Language Models (LLMs) to alleviate hallucinations and enhance response accuracy.
Retrieval-augmented generation (RAG) leverages external knowledge from rich sources to enhance large language models (LLMs) in QA, so as to mitigate hallucinations and improve response accuracy~\cite{10.1145/3726302.3729921, 10.1145/3726302.3729920, 10.1145/3726302.3729907,zhang2025ratt,zhang2025entropy,mo2026opendecoder}. 
While most investigations on RAG have focused primarily on monolingual scenario, external collections in different languages %from the same source 
can provide complementary knowledge to RAG due to their different coverage, diverse linguistic expressions and cultural features~\cite{liu-etal-2025-xrag, 10.1145/3726302.3730201,huang2026survey}. 
Therefore, a valuable practical scenario to investigate is RAG that % retrieval-augmented methods can intelligently 
exploits knowledge from multilingual collections, and in particular, from collections in other languages (that we call cross-lingual collections).
%, and has drawn much attention.
% This underscores the importance of fully exploiting cross-lingual knowledge when addressing language-specific tasks.

% Current RAG research predominantly follows a monolingual paradigm,where knowledge is retrieved from a corpus in the same language as the user query.
% However, corpora in other languages may contain information-dense sources that surpass and have higher information density than those in the query language.
% As shown in Table ~\ref{tab:cross-lingual_retrieval}, retrieving knowledge from multiple language corpora leads to further performance improvements compared to using only a monolingual knowledge base, particularly in low-resource settings. 
% This reflects the inherent knowledge disparities across languages, where specific language-specific collections often provide more comprehensive and accurate information than the native corpus.

This problem has been overlooked in the existing research. %Despite the richness of cross-lingual external knowledge, a major challenge is to effectively leverage the knowledge from non-native languages.
RAG approaches using multilingual collections simply concatenate multiple pieces of knowledge from different languages based solely on the relevance~\cite{hossain2026costefficientcrosslingualretrievalaugmentedgeneration, liu-etal-2025-xrag,mo2025survey,huang-etal-2025-boosting}. This simplistic solution leads to two main limitations.
First, all retrieved knowledge is treated equally regardless of language. This may penalize the knowledge in  %weakens the strength of 
the original language of the query, which is usually assumed to be more influential. %Knowledge from the native language loses the dominance, which leads to a degradation in performance.
Second, there is a representation gap between different languages in LLMs~\cite{qi2025sot}, so that the question representation is only weakly aligned with that of documents in other languages, leading to mismatch.
%of the retrieved knowledge and the query's native language in the representation space, where the knowledge from weaker-aligned languages causes an interference.
As a result, the existing methods trying to incorporate cross-lingual external knowledge often fail to improve effectiveness due to the potential disparities across languages.
% Existing approaches typically operate within a fixed pipeline framework, improving performance through point-wise modifications of isolated modules~\cite{}. Both the number of retrieval steps and the retrieval languages are predefined. Retrieved documents are simply concatenated as context.
% Such a "one-size-fits-all" strategy often fails to fully exploit available knowledge and can introduce noise from redundant information.

% Recent advances in reinforcement learning (RL) based post-training offer new perspectives on integrating cross-lingual knowledge. This paradigm enables LLMs to engage in dynamic interaction with search engines and perform multi-turn reasoning.
To better leverage cross-lingual knowledge for RAG, a core issue is to design an appropriate method to project cross-lingual knowledge into a unified representation space. % becomes a core issue in mitigating disparities across languages. 
This task is challenging because it is difficult to determine what cross-lingual knowledge is useful for a question and when it should be combined~\cite{10.1145/3626772.3657819, 10.1145/3626772.3657867}. Search-augmented reinforcement learning (RL) provides a possible solution to this problem~\cite{10.1145/3626772.3657760,kumar2024training,zhang2024prototypical,jin2025searchr1trainingllmsreason,zeng2026synplanresearch,mo2026agentic}. It enables LLMs to engage in dynamic interactions with search engines and perform multi-turn reasoning.
In so doing, the entire framework is optimized holistically within a search-and-reasoning framework where knowledge can be integrated at multiple stages in various forms depending on the needs.
%The paradigm represents a potential solution to optimize the cross-lingual gap and unlocks the flexibility for cross-lingual knowledge interaction with LLMs.

In this paper, we propose \textbf{\ours}, a search-augmented RL framework to address this issue. It uses Group Relative Policy Optimization (GRPO)~\cite{shao2024deepseekmathpushinglimitsmathematical} to achieve efficient and dynamic utilization of cross-lingual knowledge.
Specifically, \ours~adopts a \textit{complemental knowledge retrieval} strategy where LLMs dynamically determine the timing to invoke the search engine and perform multiple retrieval turns when encountering knowledge insufficiency.
The progressive strategy prioritizes local collection for primary knowledge in the same language during the first turn, then adaptively expands to global collections for complementary evidence in subsequent steps when needed. This strategy that prioritizes the knowledge in the same language can mitigate cross-lingual interference. %, ensuring precise knowledge acquisition.
Moreover, to address semantic gap in cross-lingual documents, we perform multilingual knowledge normalization in the process of \textit{cross-lingual knowledge integration}, bridging disparate knowledge into a unified representation space to alleviate contextual conflict, which often stems from inconsistent factual claims or linguistic biases inherent in different language corpora.
% This progressive strategy effectively mitigates cross-lingual interference, ensuring precise knowledge acquisition.
% In reinforcement learning optimization, \ours~introduce a \textit{Synchronized Cross-Lingual Policy Optimization}, which organizes semantically identical queries from different languages into a single training group rather than sampling a single language repeatedly. By transforming standard repetitive sampling into a process that encourages semantically equivalent cross-lingual multi-path reasoning, 
% queries can adapt to their respective external retrieval requirements and implicitly learn the transferability of knowledge from other languages.
% Additionally, we refine the outcome-based reward function by incorporating an n-gram-based recall score, which improves adaptability to generate responses varying lengths in different languages.
In reinforcement learning optimization, \ours~introduce a \textit{Synchronized Cross-Lingual Policy Optimization}, which enables the model to explore a single query through diverse cross-lingual ``thinking'' patterns. Instead of standard repetitive sampling in an invariable linguistic setting, each rollout within a group is guided to initiate its retrieval and reasoning in the cross-lingual manner. By transforming sampling into this multi-path cross-lingual exploration, queries can adapt to their respective external retrieval requirements and implicitly learn the transferability of knowledge from other languages.
 Experimental results demonstrate that \ours~effectively leverages cross-lingual knowledge collections as complements and improves the effectiveness of RAG with cross-lingual collections.

\section{Methodology}
\subsection{Complemental Knowledge Retrieval}

\begin{algorithm}[!t]
\SetAlFnt{\small}
\SetAlCapFnt{\small}
\SetAlCapNameFnt{\small}
\caption{Multi-Turn Dynamic Retrieval}
\label{alg:dynamic_retrieval}
\DontPrintSemicolon

\SetKwInOut{Require}{Require}
\SetKwInOut{Ensure}{Ensure}

\Require{Input query $x^{(L)}$, policy model $\pi_{\theta}$, cross-lingual search operators $\mathcal{W} = \{\omega_l\}_{l \in \mathbb{G}}$, max budget $B$.}
\Ensure{Final response $y$.}

Initialize context $C \leftarrow \{x^{(L)}\}$ and search turn count $b \leftarrow 0$\;

\While{$b < B$}{
    Generate trajectory segment $\tau \sim \pi_{\theta}(\cdot \mid C)$ until an \textit{Action Marker} is reached;\quad$C \leftarrow C \cup \{\tau\}$\;
    % $C \leftarrow C \cup \{\tau\}$\;
    
    \If{$\tau$ triggers a \textit{Search Action}}{
        Extract search query $q$ from $\tau$\;
        $b \leftarrow b + 1$\;
        
        \tcp{Complemental Knowledge Retrieval}
        % \lIf{$b=1$}{$\mathcal{S} \leftarrow \omega_L(q)$}
        % \lElseIf{$b=2$}{$\mathcal{S} \leftarrow \bigcup_{l \in \mathbb{G}} \omega_l(q)$}
        % \lElse{$\mathcal{S} \leftarrow \omega_{en}(q)$}
        \If{$b=1$}{
         $\mathcal{S} \leftarrow \omega_L(q)$ 
            % \tcp*[f]{Native Language Anchoring}
        }
        \ElseIf{$b=2$}{
            $\mathcal{S} \leftarrow \bigcup_{l \in \mathbb{G} } \omega_l(q)$ 
            % \tcp*[f]{Global Evidence Expansion}
        }
        \Else{
            $\mathcal{S} \leftarrow \omega_{en}(q)$ 
            % \tcp*[f]{High-Resource Factual Anchor}
        }
        
        % \tcp{Integrate retrieved evidence into context}
        $C \leftarrow C \cup \{\text{KnowledgeNorm}(\mathcal{S})\}$\;
    }
    \ElseIf{$\tau$ triggers a \textit{Final Response Action}}{
        \Return{Final answer extracted from $C$}\;
    }
    \Else{
        $C \leftarrow C \cup \{\text{Self-Correction Message}\}$\;
    }
}
\Return{Final response from $C$}\;
% \vspace{-3ex}
\end{algorithm}
% \vspace{-3ex}
% The rollout process for LLM response generation employs an interleaved, multi-turn retrieval mechanism designed for dynamic supplementation of cross-lingual knowledge. We define the response generation process as:

% \noindent \textbf{Multi-Turn Dynamic Retrieval.} 
As shown in Alg.~\ref{alg:dynamic_retrieval}, we formulate the LLM response generation as an adaptive, multi-turn retrieval-augmented process, where external cross-lingual evidence is incrementally incorporated only when necessary.
Following an iterative framework~\cite{jin2025searchr1trainingllmsreason}, the model alternates between internal reasoning and external search calls. The system instruction explicitly guides the model to structure its generation using special action tokens, including <think>, <search>, <information>, and <answer>.
Formally, the generation process is defined as:

$$y \sim \pi_{\theta}(\cdot \mid x^{(L)}, \mathcal{M}), \text{ with } \mathcal{M} = \bigcup_{l \in \mathbb{G}} \mathcal{S}_l$$
Here, $x^{(L)}$ denotes the user query in language $L$, and $\mathcal{S}_l$ represents evidences retrieved from collection of language $l$ using specific search operators $\{\omega_l\}_{l \in \mathbb{G}}$. 
Upon detecting search tokens, the method extracts the query and retrieves evidence $\mathcal{S}_l$, then appends it within <information> and </information> tags as additional context for the next generation step.
% The cross-lingual evidence pool $\mathcal{M}$ is not fully activated at once. 
Subsets of $\mathcal{S}_l$ are adaptively queried according to the model’s intermediate reasoning state.
% $$y \sim \pi_{\theta}(\cdot | q_L; \mathcal{R}_{multi}) = \pi_{\theta}(\cdot | q_L) \otimes \{\mathcal{R}_l\}_{l \in \mathcal{L}}$$
% where $q_L$ represents the input query in source language $L$, and $\mathcal{R}_{multi} = \{\mathcal{R}_l\}_{l \in \mathcal{L}}$ denotes the set of language-specific retrievers for each language $l$ in the multilingual collection $\mathcal{L}$. The model initially executes a <think> phase to activate its parametric memory. If internal knowledge is deemed insufficient, the model initiates a hierarchical retrieval strategy to autonomously gather supportive evidence from diverse language collections.

Specifically, the model first calls the search operator $\omega_L$ for the query language $L$ by inserting $x^{(L)}$ within <search> and </search> tags, aiming to establish a backbone using culturally and contextually relevant local evidence. 
If the model still lacks information after interacting with relevant documents, it triggers a complemental turn to search cross-lingual collections $\mathcal{M} \setminus \mathcal{S}_L$. 
In addition, the model retrieves from a high-resource language collection (e.g., English $\omega_{en}$) to provide the remaining complementary knowledge when the retrieval process is not completed after more than two turns. 
This iterative process continues until the model generates its final response within <answer> and </answer> tags, or reaches the maximum search budget $B$.
% The process continues until the model generates a final <answer> or reaches its search limit.

% \paragraph{\textbf{Turn 1: Native Language Anchoring.}} The model first invokes the retriever $\mathcal{R}_L$ corresponding to the query's native language $L$. This step aims to capture foundational knowledge and local context, providing a culturally and contextually relevant baseline for the response.

% \paragraph{\textbf{Turn 2: Global Knowledge Refinement.}} After evaluating the native results within a subsequent <think> block, if the model identifies persistent information deficits, it triggers a second retrieval turn across global collections. 
% % The model is explicitly guided to reconstruct the retrieved global evidence into brief fact-statements in the target language within the <think> tags. 
% This process refines the initial retrieval results, aiming to utilize information-dense evidence from other languages to fill gaps in foundational knowledge.

% \paragraph{\textbf{Turn $\ge$ 3: High-Resource Factual Anchoring.}} For any further turns, the model defaults to a high-resource retriever (e.g., English $\mathcal{R}_{en}$) to serve as a stable factual anchor until a final <answer> is generated or the action budget is exhausted.

% Our hierarchical strategy enables the model to autonomously assess knowledge adequacy at each step, effectively reducing redundancy and avoiding potential conflicts.
In contrast to strategies that simply concatenate knowledge from all languages, our multi-turn retrieval strategy leverages query language evidence as a foundational anchor, maintaining the dominance of the original context. 
By introducing cross-lingual knowledge as a selective supplement to fill specific factual gaps, this approach effectively prevents redundant information from creating noise or interfering with the reasoning process.

\subsection{Cross-lingual Knowledge Integration}
To fully exploit the potential of knowledge collections across languages, we propose a cross-lingual knowledge integration mechanism aimed at harmonizing multiple knowledge and bridging semantic gaps.

\noindent \textbf{Multilingual Knowledge Normalization.} To address the semantic gap prevalent in non-native documents, we first perform the multilingual knowledge normalization to transfer the representation from other languages to the original language. 
Given a set of retrieved global documents $\mathcal{D}_{global}$ in various languages, we apply a translation operator $\mathcal{T}$ to map each document $d \in \mathcal{D}_{global}$ into a unified semantic space aligned with the query language $L$:$$\hat{\mathcal{D}}_{global} = \{ \mathcal{T}(d, L) \mid d \in \mathcal{D}_{global} \}$$where $\hat{\mathcal{D}}_{global}$ represents the normalized global evidence, facilitating the seamless fusion of multilingual information.

\noindent \textbf{Fact-based Evidence Reconstruction.}
Building upon the normalization, the model is explicitly guided to reconstruct the normalized global evidence into brief fact-statements $\mathcal{F}$ within the <think> tags.
Steered by the specialized prompt $\mathbb{P}_{recon}$, the model is asked to contrast the newly acquired cross-lingual evidence against the first-round results in the original language to resolve any factual conflicts or inherent biases.
Formally, the reasoning process is defined as:$$\mathcal{F} \sim \pi_\theta (\cdot \mid x^{(L)}, \mathcal{D}_{native}, \hat{\mathcal{D}}_{global}; \mathbb{P}_{recon})$$
where $\mathcal{D}_{native}$ is the original evidence acquired in the first retrieval turn, and $\mathbb{P}_{recon}$ denotes the specialized prompt for fact reconstruction. 
% Rather than a simple concatenation of raw documents, these reconstructed global facts $\mathcal{F}$ serve as information-dense supplementary signals to refine the native knowledge. 
These reconstructed global facts $\mathcal{F}$ are not a simple concatenation of the original documents, but rather serve as information-dense complementary signals to refine the original knowledge.

Finally, the policy model $\pi_\theta$ synthesizes the predictive response $y$ by integrating the foundational original evidence with the reconstructed cross-lingual facts:$$y \sim \pi_\theta (\cdot \mid x^{(L)}, \mathcal{D}_{native}, \mathcal{F})$$
The reconstruction leverages cross-lingual complementarity to bridge core information gaps while preserving the centrality of the context in the original language of the query, thereby minimizing interference due to knowledge disparities.

% 定义颜色
\definecolor{colorMethods}{RGB}{255, 255, 255}
\definecolor{colorGroup1}{RGB}{226, 243, 246}
\definecolor{colorGroup2}{RGB}{238, 228, 241}
\definecolor{colorAve}{RGB}{250, 235, 215}
\definecolor{slate}{gray}{0.5}

% 定义简写宏：格式为 "Fem / C3"
% #1 是 Fem (加粗), #2 是 c3recall (灰色小字)
\newcommand{\res}[2]{#1\,\textcolor{slate}{/}\,#2}
% 定义我们自己的方法名

\begin{table*}[!t]
    \centering
    \small
    \renewcommand{\arraystretch}{1.15}
    \setlength{\tabcolsep}{3.5pt} 
    \caption{Performance on MKQA and English benchmarks. Background colors indicate \colorbox{colorGroup1}{training-free} and \colorbox{colorGroup2}{post-training} paradigms.}
    % \vspace{-3ex}
    \label{tab:combined_results}
    \resizebox{0.95\textwidth}{!}{
    \begin{tabular}{ l ccccc  cccc }
        \toprule
        \multirow{2}{*}{\textbf{Methods}} & \multicolumn{5}{c}{\textbf{MKQA (\textit{cross-lingual}, fEM / c3Recall)}} & \multicolumn{4}{c}{\textbf{English Datasets (\textit{monolingual}, fEM / c3Recall)}} \\
        \cmidrule(lr){2-6} \cmidrule(lr){7-10}
        & \textbf{\textproc{En}} & \textbf{\textproc{Fr}} & \textbf{\textproc{Th}} & \textbf{\textproc{Ar}} & \textbf{Ave} & \textbf{\textproc{HotpotQA}} & \textbf{\textproc{PopQA}} & \textbf{\textproc{2Wiki}} & \textbf{\textbf{Ave}} \\
        \midrule

        % --- Qwen2.5-3B-Instruct ---
        \multicolumn{10}{c}{\cellcolor{colorMethods}\textbf{\texttt{Qwen2.5-3B-Instruct}}} \\
        \midrule
        \rowcolor{colorGroup1}Direct Inference & \res{41.11}{51.80} & \res{29.21}{40.04} & \res{11.43}{18.75} & \res{7.83}{17.60} & \res{22.40}{32.05} & \res{19.60}{35.97} & \res{18.50}{30.50} & \res{36.60}{44.79} & \res{24.90}{37.09} \\
        \rowcolor{colorGroup1}IRCoT            & \res{38.83}{51.49} & \res{39.88}{49.29} & \res{15.46}{21.13} & \res{12.46}{20.61} & \res{26.66}{35.63} & \res{35.42}{42.70} & \res{52.68}{60.44} & \res{34.17}{44.58} & \res{40.75}{49.24} \\
        \rowcolor{colorGroup1}Search-o1        & \res{41.66}{53.23} & \res{38.62}{51.05} & \res{15.25}{20.35} & \res{13.67}{21.81} & \res{27.30}{36.61} & \res{30.44}{41.68} & \res{54.04}{62.50} & \res{33.12}{45.65} & \res{39.20}{49.94} \\
        \rowcolor{colorGroup1}RAG              & \res{52.08}{60.73} & \res{40.72}{50.20} & \res{14.02}{19.04} & \res{11.59}{17.42} & \res{29.60}{36.85} & \res{31.50}{43.84} & \res{50.10}{59.47} & \res{38.60}{45.90} & \res{40.07}{49.74} \\
        \rowcolor{colorGroup2}SFT              & \res{56.85}{63.60} & \res{\underline{48.00}}{56.56} & \res{16.91}{29.06} & \res{16.84}{28.23} & \res{34.65}{44.36} & \res{\textbf{39.20}}{50.79} & \res{59.50}{65.18} & \res{38.90}{45.90} & \res{45.87}{53.96} \\
        \rowcolor{colorGroup2}Search-R1        & \res{\underline{63.38}}{\underline{77.31}} & \res{44.73}{\underline{61.35}} & \res{\underline{22.36}}{\underline{33.95}} & \res{\underline{16.89}}{\underline{39.49}} & \res{\underline{36.82}}{\underline{53.03}} & \res{35.35}{\underline{62.65}} & \res{\underline{59.96}}{\underline{78.39}} & \res{\underline{52.93}}{\underline{65.76}} & \res{\underline{49.41}}{\underline{68.93}} \\
        \rowcolor{colorAve}Ours & \res{\textbf{67.68}}{\textbf{83.22}} & \res{\textbf{54.00}}{\textbf{72.39}} & \res{\textbf{24.41}}{\textbf{42.26}} & \res{\textbf{17.48}}{\textbf{47.87}} & \res{\textbf{40.89}}{\textbf{61.44}} & \res{\underline{38.67}}{\textbf{66.88}} & \res{\textbf{65.04}}{\textbf{83.46}} & \res{\textbf{55.26}}{\textbf{70.42}} & \res{\textbf{52.99}}{\textbf{73.59}} \\

        \midrule
        % --- Qwen2.5-7B-Instruct ---
        \multicolumn{10}{c}{\cellcolor{colorMethods}\textbf{\texttt{Qwen2.5-7B-Instruct}}} \\
        \midrule
        \rowcolor{colorGroup1}Direct Inference & \res{49.96}{60.00} & \res{35.79}{46.86} & \res{14.02}{20.75} & \res{10.73}{19.69} & \res{27.63}{36.83} & \res{26.50}{43.83} & \res{24.90}{37.06} & \res{33.80}{41.59} & \res{28.40}{40.83} \\
        \rowcolor{colorGroup1}IRCoT            & \res{40.88}{53.67} & \res{41.52}{51.44} & \res{18.06}{23.65} & \res{15.28}{23.84} & \res{28.94}{38.15} & \res{40.84}{47.67} & \res{31.43}{50.22} & \res{33.34}{44.85} & \res{35.20}{47.58} \\
        \rowcolor{colorGroup1}Search-o1        & \res{44.24}{56.06} & \res{41.48}{53.57} & \res{18.54}{23.65} & \res{15.16}{24.53} & \res{29.86}{39.45} & \res{36.32}{47.62} & \res{30.48}{47.06} & \res{33.92}{46.88} & \res{33.57}{47.19} \\
        \rowcolor{colorGroup1}RAG              & \res{55.05}{64.12} & \res{44.64}{53.53} & \res{16.76}{22.55} & \res{16.37}{24.43} & \res{33.21}{41.16} & \res{37.60}{50.34} & \res{10.00}{55.00} & \res{39.70}{46.69} & \res{29.10}{50.68} \\
        \rowcolor{colorGroup2}SFT              & \res{59.91}{66.86} & \res{51.92}{59.53} & \res{23.26}{34.69} & \res{\underline{19.81}}{29.25} & \res{38.73}{47.58} & \res{\underline{43.00}}{54.61} & \res{\textbf{62.60}}{68.40} & \res{39.80}{46.25} & \res{48.47}{56.42} \\
        \rowcolor{colorGroup2}Search-R1        & \res{\underline{69.43}}{\underline{79.82}} & \res{\underline{57.91}}{\underline{70.36}} & \res{\underline{25.29}}{\underline{37.13}} & \res{18.65}{\underline{34.68}} & \res{\underline{42.82}}{\underline{55.50}} & \res{\textbf{44.14}}{\underline{67.17}} & \res{58.20}{\underline{70.96}} & \res{\underline{56.05}}{\underline{66.08}} & \res{\underline{52.80}}{\underline{68.07}} \\
        \rowcolor{colorAve}Ours & \res{\textbf{72.07}}{\textbf{82.68}} & \res{\textbf{59.67}}{\textbf{74.04}} & \res{\textbf{27.83}}{\textbf{45.98}} & \res{\textbf{24.12}}{\textbf{41.63}} & \res{\textbf{45.92}}{\textbf{61.08}} & \res{\textbf{44.14}}{\textbf{67.70}} & \res{\underline{61.33}}{\textbf{78.04}} & \res{\textbf{61.13}}{\textbf{73.64}} & \res{\textbf{55.53}}{\textbf{73.13}} \\
        \bottomrule
    \end{tabular}
    }
% \vspace{-2ex}    
\end{table*}

\subsection{Synchronized Cross-Lingual Policy Optimization}
% To accommodate multilingual scenarios, we implement a language-coupled rollout and optimization mechanism within the GRPO framework. This approach encourages the model to learn transferable "thinking habits" by evaluating reasoning trajectories across a semantically equivalent multilingual group.

To handle the complexities of cross-lingual environment, we introduce a synchronized optimization strategy built upon  GRPO. This design aims to train the model to develop universal reasoning capability for cross-lingual scenarios by comparing the thinking paths for the same question in different languages.

\noindent \textbf{Parallel Trajectory Sampling.}
% To foster a unified understanding across languages, we construct a cluster of semantically equivalent queries $\mathcal{X} = \{x_1, x_2, \dots, x_k\}$ for a given source input $x$. Each variation $x_i$ is generated through cross-lingual thinking strategies to ensure that the core intent remains invariant while the linguistic form changes. 
To foster a unified understanding across languages, we explore the source query $x$ through a set of cross-lingual thinking modes $\mathcal{L} = \{l_1, l_2, \dots, l_G\}$, where each mode $l_i$ guides the model to initiate its reasoning and retrieval within a specific language context. This approach ensures that the core semantic intent remains unchanged while the reasoning trajectory is conditioned on  different languages.
During the rollout phase, we perform $G$ independent samplings for the query $x$. For the $i$-th sampling ($i \in \{1, \dots, G\}$), the model generates a reasoning response $r_i$ under the guidance of the cross-lingual thinking mode $l_i$:
% $$r_i \sim \pi_{\theta}(\cdot \mid x_i, \mathcal{E}_i)$$
$$r_i \sim \pi_{\theta}(\cdot \mid x, \mathcal{E}_i, l_i)$$
where $\mathcal{E}_i$ represents the evidence retrieved from the global collection in $i$-th sampling. We optimize the policy $\pi_{\theta}$ by maximizing:

$$\small
 \begin{aligned}\mathcal{L}_{\text{CLPO}}(\theta) = &\mathbb{E}_{\{\tau_i\}_{i=1}^G} \Bigg[ \frac{1}{G} \sum_{i=1}^G \frac{1}{|\mathcal{T}_i|} \sum_{t \in \mathcal{T}_i} \\
 &\min \left( \rho_{i,t}(\theta) \tilde{A}_{i,t}^{cross},  \text{clip}(\rho_{i,t}(\theta), 1 - \delta, 1 + \delta) \tilde{A}_{i,t}^{cross} \right) \\
&- \lambda \cdot \frac{1}{G} \sum_{i=1}^G \text{KL}(\pi_{\theta}(\cdot \mid x^{(l_i)}) \parallel p_{\text{ref}}(\cdot \mid x^{(l_i)})) \Bigg] \end{aligned}$$
where $\rho_{i,t}(\theta)$ denotes the importance ratio between the updated policy and the behavior policy at step $t$, $\tilde{A}_{i,t}^{cross}$ is the advantage baseline computed using cross-lingual sampling group, and $\mathcal{T}_i$ indexes the time steps associated with reasoning, retrieval invocation, and fact synthesis actions. 
% Unlike standard GRPO which gather responses on a single-query group, we construct more discriminative thinking paths by leveraging a cross-lingual paradigm, allowing the model to explore multiple linguistic paths for the same underlying intent. 
% Instead of sampling multiple responses for a single query as in standard GRPO, we obtain distinct thinking paths by sampling from a group of semantically equivalent queries across different languages. 
The standard GRPO follows a uniform sampling procedure where multiple responses are generated from a given query, resulting in relatively homogeneous reasoning. In contrast,  we guide the query to think in different languages, sampling across a group of semantically equivalent queries.
These samples are complementary because they represent diverse reasoning and retrieval trajectories triggered by the unique knowledge in each language.

% By exposing the policy to rollouts conditioned on semantically equivalent queries across multiple languages, the model is trained to process newly introduced knowledge under diverse linguistic realizations. 
By exposing the policy to rollouts conditioned on the cross-lingual thinking modes, the model is trained to process the trajectories in various expressions of the same semantics.
% This rollout setting encourages the policy to align reasoning behaviors beyond surface-level language forms, thereby mitigating representational discrepancies across languages and ultimately enhancing cross-lingual transferability.
The cross-lingual rollout can acquire diverse training samples.
This ensures that the model learns a broader range of semantic retrieval and generation relationships, thereby enhancing its cross-lingual transferability.
% The coefficients $\delta$ and $\lambda$ control policy update stability and regularization strength, respectively.

\noindent \textbf{N-gram Outcome Reward.}
Rule-based outcome rewards based on exact string matching, as adopted in prior studies such as Search-R1, are ill-suited for cross-lingual settings due to their sensitivity to surface-level lexical variation.
To address this limitation, we employ a character-level similarity metric based on 3-gram recall to evaluate answer correctness across languages.

Formally, the outcome score for trajectory $\tau_i$ is defined as:
$$r_{\text{out}}(i) = \text{Char-3-Recall}(\hat{y}_i, y^{\star})$$
where $\hat{y}_i$ denotes the extracted answer span and $y^{\star}$ is the reference answer. This metric provides a smooth and dense reward signal that mitigates abrupt reward discontinuities, reducing the likelihood of unstable gradient updates caused by extreme importance weight amplification.
% Previous methods~(e.g., Search-R1) adopt the rule-based outcome reward system using exact matching (EM), but this discrete approach is suboptimal for multilingual tasks where lexical diversity is prevalent.
% As word-level matching fails to capture such lexical similarities, we replace it with Character 3-gram Recall ($r_{\text{ans}}$) ~\cite{chirkova2024retrieval} to assess answer correctness across multilingual generations:$$r_{\text{ans}}(i) = \text{c3Recall}(\hat{a}_i, a_{\text{gold}})$$where $\hat{a}_i$ is the extracted answer and $a_{\text{gold}}$ is the ground truth. Unlike EM, $r_{\text{ans}}$ provides a dense reward signal, which slows down the entry into ``high-gradient danger zones'' by stabilizing importance ratios, preventing the advantages from being amplified by extreme reward disparities.

\section{Experiments}
\subsection{Experimental Settings}

% \noindent \textit{Datasets and Evaluation Metrics.} 
We evaluate our method on both multilingual knowledge-intensive benchmark MKQA~\cite{longpre-etal-2021-mkqa} in Latin languages (\textproc{En}, \textproc{Fr}) and non-Latin languages (\textproc{Th}, \textproc{Ar}) and representative English benchmarks including HotpotQA~\cite{yang-etal-2018-hotpotqa}, PopQA~\cite{popqa-etal-2023-trust} and 2WikimultihopQA~\cite{ho-etal-2020-2wiki}.
For all benchmarks, we utilize a set of Wikimedia\_dump\footnote{\url{https://huggingface.co/datasets/wikimedia/wikipedia}} from different languages (French, English, Arabic, and Thai) as the cross-lingual collections.
All methods in the experiment are allowed to utilize the full collections for a fair comparison.
Retriever and Translator are selected as multilingual E5~\cite{wang2024multilingual} and nllb-200-distilled-600M~\cite{nllbteam2022languageleftbehindscaling}, respectively. 
To ensure a comprehensive evaluation, we evaluate against: (1) Direct Inference without Retrieval; (2) Inference with Retrieval, including naive RAG~\cite{lewis2020retrieval}, IRCoT~\cite{trivedi2023interleaving} and Search-o1~\cite{li2025search}; and (3) Fine-Tuning Methods such as Supervised fine-tuning~(SFT)~\cite{chung2024scaling} and a cross-lingual reproduction of Search-R1~\cite{jin2025searchr1trainingllmsreason}.
We employ Qwen2.5-3B-Instruct and Qwen2.5-7B-Instruct~\cite{qwen2025qwen25technicalreport} as the primary backbones for our experiments. 
% For knowledge retrieval, we use multilingual E5~\cite{wang2024multilingual} to search in Wikimedia\_dump\footnote{\url{https://huggingface.co/datasets/wikimedia/wikipedia}}, with nllb-200-distilled-600M\footnote{\url{https://huggingface.co/facebook/nllb-200-distilled-600M}} acting as the translator. 
For evaluation, we follow ~\citet{chirkova2024retrieval} and use both flexible exact match (fEM) and recall on character 3-gram level (c3Recall)~\cite{schick2023toolformer}.

\begin{table}[!t]
    \centering
    \small
    % \resizebox{0.48\textwidth}{!}{
    \caption{Average results of \ours~without each module based on \texttt{\textbf{Qwen2.5-3B-Instruct}} for MKQA.}
    \label{tab:ablation_results}
    %\vspace{-3ex}
    \begin{tabular}{lccc}
        \toprule
         \textbf{ } & \textbf{fEM} & \textbf{c3Recall}  \\
        \midrule
        \rowcolor{gray!20}Our Full Framework & 40.89 & 61.44 \\
        \midrule
        % w/ Monolingual Knowledge Retrieval & 40.58 & 59.67 \\
        w/o Cross-lingual Knowledge Integration &  38.46 & 55.68 \\
        % \midrule
        w/o Cross-lingual Policy Optimization &  38.40 & 56.44  \\
        w/o c3Recall Reward &  28.85 & 33.49 \\
        % \midrule
        % w/. GRPO & 30.0 & 44.9 & - \\
        % Replace by PPO & 15.5 & 21.7 & 72.3\\
        % \rowcolor{gray!20}\ours~(Ours) & - & - & - \\
        \bottomrule
    \end{tabular}%
     % }
     % \vspace{-3ex}
\end{table}

\subsection{Results}
\noindent \textbf{Overall.}
% As shown in Table~\ref{tab:main_result_mkqa}, we investigate the performance of \ours~ compared with other methods on MKQA.
% The results demonstrate that our proposed method (\ours) outperforms several strong baselines in terms of all evaluation metrics for various LLMs.
% Although the search-augmented RL method~(Search-R1) facilitates the knowledge transfer in the cross-lingual settings compared to other post-training methods such as SFT, the parameters are still hard to optimize, and its performance has not been significantly improved. 
The left half of Table~\ref{tab:combined_results} shows the performance on MKQA, where \ours consistently surpasses strong baselines on LLMs of different sizes with respect to fEM and c3Recall. 
While Search-R1 demonstrates robust cross-lingual transferability, it remains hindered by optimization difficulties and marginal performance gains.
In addition, as shown in the right half of Table~\ref{tab:combined_results}, the results on English benchmarks demonstrate that cross-lingual resources generally make positive impact with \ours, while they may degrade the performance with other methods due to knowledge disparities.  This phenomenon is particularly pronounced for the resource-rich language - English.
In fact, existing methods simply concatenates original and cross-lingual evidence. It is easy for cross-lingual knowledge to introduce noise that interferes with that of the original language. %through direct context injection. 
%Therefore, cross-lingual knowledge remains underutilized and even introduces contextual interference with the original information. 
%We address these limitations by incorporating complemental knowledge retrieval and cross-lingual knowledge integration. 
% Through our cross-lingual sampling mechanism into the RL process, our approach facilitates effective knowledge transfer, enabling the model to better supplement and refine original knowledge using cross-lingual knowledge.
% We address this issue by employing a multi-turn retrieval strategy that preserves native knowledge as a foundational anchor to prevent cross-lingual interference, while cross-lingual knowledge integration transforms external evidence into harmonized fact-statements for better comprehension; furthermore, our cross-lingual sampling mechanism optimizes the policy through diverse linguistic thinking patterns, making the model more proficient at utilizing cross-lingual knowledge.
Our method retains the dominance of the original language while introducing cross-lingual thinking patterns during sampling to generate trajectories with linguistic diversity. 
This expands the state space for cross-lingual retrieval and generation, allowing the model to better leverage cross-lingual knowledge and brige the semantic gap, compared with Search-R1.

\noindent \textbf{Ablation Studies.} 
% Table~\ref{tab:ablation_results} investigates the impact of different modules in \ours.
% The results demonstrate that the removal of each module leads consistently to a performance degradation, thereby validating the utility of complemental knowledge retrieval, cross-lingual knowledge integration, and synchronized cross-lingual policy optimization.
% In particular, cross-lingual alignment optimization makes the largest impact. 
% %mitigates knowledge conflicts and has a greater impact.
% The complementary knowledge retrieved from cross-lingual knowledge collections further enhances performance compared to being invariably extracted from the query language.
% Additionally, we refine the outcome-based reward function by incorporating an n-gram-based recall score, which improves adaptability to generate responses varying lengths in different languages.
Table~\ref{tab:ablation_results} investigates the impact of different modules in \ours. The results demonstrate that the removal of each module leads consistently to a performance degradation. 
% Specifically, the complemental knowledge retrieval provides a structurally stable retrieval backbone for the following steps. The lightweight modifications to the retrieval pipeline yields measurable gains, indicating that structured retrieval design contributes substantially to robustness. Furthermore, synchronized cross-lingual policy optimization increases exposure to cross-lingual evidence patterns, leading to better utilization of cross-lingual knowledge. Moreover, cross-lingual knowledge integration produces the most significant improvement, shows that the critical factor lies in how heterogeneous knowledge is normalized and reconstructed.
Specifically, cross-lingual knowledge integration and cross-lingual policy optimization have greater impacts. 
This indicates that it is more crucial to eliminate representation discrepancies when injecting cross-lingual knowledge, as LLMs can better understand such knowledge effectively for the query language.
% Additionally, we refine the outcome-based reward function by incorporating an n-gram-based recall score, which improves adaptability to generate responses varying lengths in differentlanguages.
Compared to the outcome-based reward function for exact matching, the n-gram score broadens the scope of rewards, allowing for greater flexibility in generated answers, particularly when the length of outputs varies significantly the for cross-lingual scenario.

\subsection{Analysis}
\begin{figure}[t]
    \centering

    \begin{subfigure}[b]{0.8\linewidth}
        \centering
        \includegraphics[width=\linewidth]{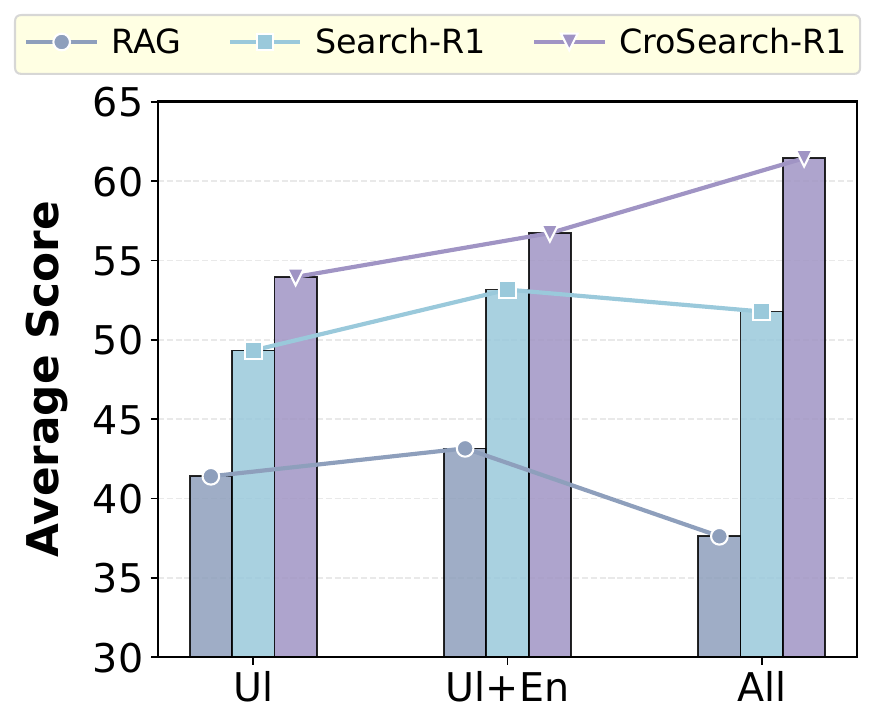}
        % \vspace{-2ex}
        \caption{Number of language collections.}
        \label{fig:analysis_1}
    \end{subfigure}

    % \vspace{1ex} % 控制上下间距（可调）

    \begin{subfigure}[b]{0.9\linewidth}
        \centering
        \includegraphics[width=\linewidth]{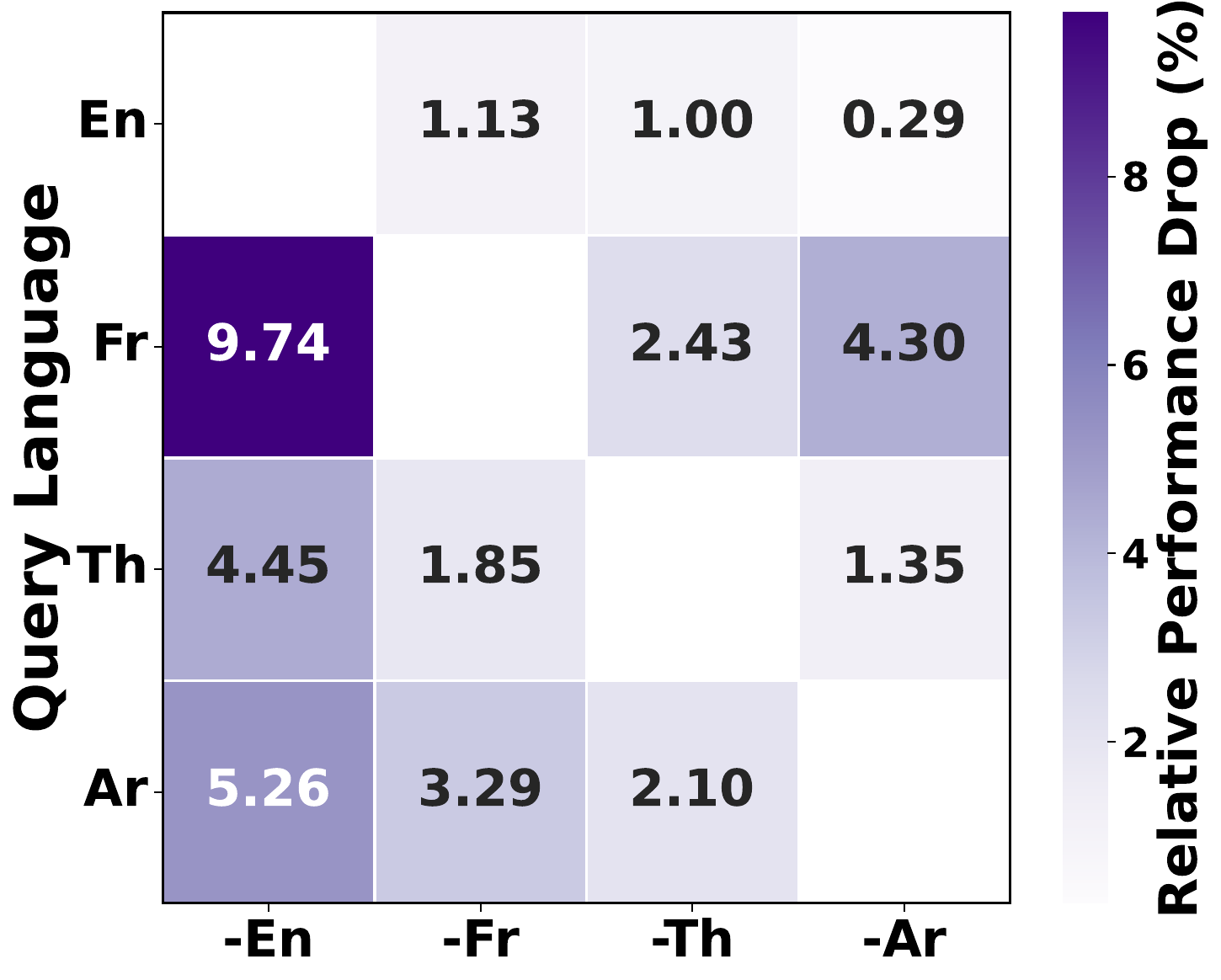}
        % \vspace{-2ex}
        \caption{Type of language collections.}
        \label{fig:analysis_2}
    \end{subfigure}

    % \vspace{-2ex}
    \caption{Performance variation caused by language collections. Ul of (a) indicates the query language and the value within each block in (b) represents the degradation when document collections in another language (X-axis) are removed.}
    % \vspace{-3ex}
    \label{fig:analysis}
\end{figure}

\noindent \textbf{Scalability of Increasing Language Counts.}
% As shown in Figure~\ref{fig:analysis_1}, we investigate the impact of increasing language diversity in the retrieval collections on different RAG methods. 
% The results show that only our method exhibits a steady improvement, whereas other approaches suffer a sharp performance degradation when the number of languages in the retrieval collections further increases. 
Figure~\ref{fig:analysis_1} illustrates the variation in retrieval performance with increasing language diversity. 
% While rival RAG methods decline significantly as more languages are added, our approach alone maintains a consistent upward trend.
In contrast to other RAG methods, whose performance significantly declines with the increase in the number of languages, our method consistently maintains an upward trend.
% The degradation of RAG baselines is primarily caused by inter-lingual interference, where conflicting or redundant facts from heterogeneous sources overwhelm the model's reasoning.
This performance degradation is primarily caused by the interference of cross-lingual knowledge, where the potential disparities in knowledge across languages may confuse the model. %'s reasoning ability.
This situation corresponds to the research question we identified and our method addresses this by employing a complemental injection strategy that selectively integrates cross-lingual evidence when necessary, ensuring that the original context remains the dominant anchor for the final response.
% Our method explicitly addresses this issue through a hierarchical injection strategy following cross-lingual knowledge integration.
% Our method addresses this by employing a hierarchical injection strategy that selectively integrates global evidence when necessary, ensuring that the native context remains the authoritative anchor for the final response.

\noindent \textbf{Effect of Language-specific Collections.}
As shown in Figure~\ref{fig:analysis_2}, we investigate the impact of language-specific collections for different query languages.
The results show that the exclusion of English knowledge leads to a substantial performance degradation across all other languages and the performance on English queries remains robust regardless of the removal of any other language collections.
It reveals the high density of valuable information within the English collection, which is directly proportional to its predominant corpus scale.
Interestingly, we identify significant mutual knowledge synergy between specific language pairs. For instance, French and Arabic exhibit high reciprocal utility, where external knowledge from one language significantly enhances the response quality of the other.
However, the synergy between Thai and Arabic is relatively limited, with each language providing only marginal gains for 
queries in the other.
% \begin{figure}
%     \centering
%     \includegraphics[width=0.5\linewidth]{cross_pic2.png}
%     \caption{Analysis 1}
%     \label{fig:language_benefit}
% \end{figure}

% \begin{figure}[t] % 不带星号，限制在单栏内
%     \centering
%     \begin{subfigure}[b]{0.48\linewidth}
%         \centering
%         \includegraphics[width=\linewidth]{Scaling_Benchmark_StyleV2.pdf}
%         \vspace{-2ex}
%         \caption{Number of language collections.}
%         \label{fig:analysis_1}

%     \end{subfigure}
%     \hfill
%     \begin{subfigure}[b]{0.48\linewidth}
%         \centering
%         \includegraphics[width=\linewidth]{cross_pic2.pdf}
%         \vspace{-2ex}
%         \caption{Type of language collections.}
%         \label{fig:analysis_2}

%     \end{subfigure}
%     \vspace{-2ex}
%     \caption{Performance variation caused by language collections. Ul of (a) indicates the query language and the value within each block in (b) represents the degradation when document collections in another language (X-axis) are removed.}
%     \vspace{-3ex}
%     \label{fig:analysis}

% \end{figure}

 \section{Related Work}
Retrieval-Augmented Generation~\cite{lewis2020retrieval,su2025parametric,mo2025uniconv} improves question answering by incorporating external knowledge during content generation, using additional context to achieve better accuracy for LLMs~\cite{wu2024languagesequalinsightsmultilingual,su2025parametriclora,zhang2026starpo}. In cross-lingual settings, relevant knowledge is often distributed across documents in different languages, and direct integration may introduce noise or lead to conflicting information, which harms generation quality. Existing cross-lingual RAG methods commonly adopt a translation-based strategy~\cite{ranaldi2025multilingualretrievalaugmentedgenerationknowledgeintensive, chirkova2024retrieval, 10756977, ahmad2024enhancingmultilingualinformationretrieval, park-lee-2025-investigating}, translating documents into a single language before retrieval and generation. This approach reduces language gaps to some extent, but noise introduced by translation errors may lead to the loss of key information. Other methods focus on modifying the knowledge injection process~\cite{10594504, hossain2026costefficient} or the reasoning strategy~\cite{kumar-etal-2025-bridging, ranaldi-etal-2025-improving-multilingual, li2025languagedriftmultilingualretrievalaugmented}. However, modifying knowledge injection often disrupts semantic integrity, while complex reasoning strategies remain prone to "language drift"~\cite{li2025languagedriftmultilingualretrievalaugmented} and hallucination without robust filtering of cross-lingual evidence. 

In contrast to these methods, we fully leverage cross-lingual knowledge through Cross-lingual Knowledge Integration and multi-path sampling optimization, while employing a progressive search strategy with an iterative RL method to mitigate the interference of cross-lingual knowledge from different languages.

\section{Conclusions}
\label{sec:subsection}

In this paper, we present CroSearch-R1, a search-augmented RL framework designed to effectively leverage cross-lingual knowledge for RAG. 
By integrating a multi-turn retrieval and cross-lingual knowledge integration strategy into a synchronized cross-lingual policy optimization framework, the approach enables dynamic knowledge complement and multi-path reasoning transfer across languages.
Experimental results demonstrate that \ours~effectively exploits cross-lingual knowledge retrieval and significantly bridges the gap between language representations.

\begin{acks}
This work is supported by the Fundamental Research Funds for the Central Universities of China under Grant 2024JBGP008 and the National Natural Science Foundation of China (No. 62406018, 62476023, 61976016, 62376019), and the authors would like to thank the anonymous reviewers for their valuable comments and suggestions to improve this paper.
\end{acks}

\bibliographystyle{ACM-Reference-Format}
\bibliography{main}

@String{Computing = "Computing" }

@String{Springer = "Springer-Verlag" }

@inproceedings{chirkova2024retrieval,
  title={Retrieval-augmented generation in multilingual settings},
  author={Chirkova, Nadezhda and Rau, David and D{\'e}jean, Herv{\'e} and Formal, Thibault and Clinchant, St{\'e}phane and Nikoulina, Vassilina},
  booktitle={Proceedings of the 1st Workshop on Towards Knowledgeable Language Models (KnowLLM 2024)},
  pages={177--188},
  year={2024}
}

@article{schick2023toolformer,
  title={Toolformer: Language models can teach themselves to use tools},
  author={Schick, Timo and Dwivedi-Yu, Jane and Dess{\`\i}, Roberto and Raileanu, Roberta and Lomeli, Maria and Hambro, Eric and Zettlemoyer, Luke and Cancedda, Nicola and Scialom, Thomas},
  journal={Advances in Neural Information Processing Systems},
  volume={36},
  pages={68539--68551},
  year={2023}
}

@misc{jin2025searchr1trainingllmsreason,
      title={Search-R1: Training LLMs to Reason and Leverage Search Engines with Reinforcement Learning}, 
      author={Bowen Jin and Hansi Zeng and Zhenrui Yue and Jinsung Yoon and Sercan Arik and Dong Wang and Hamed Zamani and Jiawei Han},
      year={2025},
      eprint={2503.09516},
      archivePrefix={arXiv},
      primaryClass={cs.CL},
      url={https://arxiv.org/abs/2503.09516}, 
}

@article{kumar2024training,
  title={Training language models to self-correct via reinforcement learning},
  author={Kumar, Aviral and Zhuang, Vincent and Agarwal, Rishabh and Su, Yi and Co-Reyes, John D and Singh, Avi and Baumli, Kate and Iqbal, Shariq and Bishop, Colton and Roelofs, Rebecca and others},
  journal={arXiv preprint arXiv:2409.12917},
  year={2024}
}

@article{longpre-etal-2021-mkqa,
    title = "{MKQA}: A Linguistically Diverse Benchmark for Multilingual Open Domain Question Answering",
    author = "Longpre, Shayne  and
      Lu, Yi  and
      Daiber, Joachim",
    editor = "Roark, Brian  and
      Nenkova, Ani",
    journal = "Transactions of the Association for Computational Linguistics",
    volume = "9",
    year = "2021",
    address = "Cambridge, MA",
    publisher = "MIT Press",
    url = "https://aclanthology.org/2021.tacl-1.82/",
    doi = "10.1162/tacl_a_00433",
    pages = "1389--1406"
}

@inproceedings{kumar-etal-2025-bridging,
    title = "Bridging the Language Gap: Dynamic Learning Strategies for Improving Multilingual Performance in {LLM}s",
    author = "Kumar, Somnath  and
      Balloli, Vaibhav  and
      Ranjit, Mercy  and
      Ahuja, Kabir  and
      Sitaram, Sunayana  and
      Bali, Kalika  and
      Ganu, Tanuja  and
      Nambi, Akshay",
    editor = "Rambow, Owen  and
      Wanner, Leo  and
      Apidianaki, Marianna  and
      Al-Khalifa, Hend  and
      Eugenio, Barbara Di  and
      Schockaert, Steven",
    booktitle = "Proceedings of the 31st International Conference on Computational Linguistics",
    month = jan,
    year = "2025",
    address = "Abu Dhabi, UAE",
    publisher = "Association for Computational Linguistics",
    url = "https://aclanthology.org/2025.coling-main.619/",
    pages = "9209--9223"
}

@inproceedings{park-lee-2025-investigating,
    title = "Investigating Language Preference of Multilingual {RAG} Systems",
    author = "Park, Jeonghyun  and
      Lee, Hwanhee",
    editor = "Che, Wanxiang  and
      Nabende, Joyce  and
      Shutova, Ekaterina  and
      Pilehvar, Mohammad Taher",
    booktitle = "Findings of the Association for Computational Linguistics: ACL 2025",
    month = jul,
    year = "2025",
    address = "Vienna, Austria",
    publisher = "Association for Computational Linguistics",
    url = "https://aclanthology.org/2025.findings-acl.295/",
    doi = "10.18653/v1/2025.findings-acl.295",
    pages = "5647--5675",
    ISBN = "979-8-89176-256-5"
}

@inproceedings{zhang2025entropy,
  title={Entropy-based exploration conduction for multi-step reasoning},
  author={Zhang, Jinghan and Wang, Xiting and Mo, Fengran and Zhou, Yeyang and Gao, Wanfu and Liu, Kunpeng},
  booktitle={Findings of the Association for Computational Linguistics: ACL 2025},
  pages={3895--3906},
  year={2025}
}

@article{zhang2026starpo,
      title={StaRPO: Stability-Augmented Reinforcement Policy Optimization}, 
      author={Jinghan Zhang and Fengran Mo and Tharindu Cyril Weerasooriya and Ruimin Dai and Xiaoyan Han and Yanjie Fu and Dakuo Wang and Kunpeng Liu},
      year={2026},
      journal={arXiv preprint arXiv:2604.08905},
}

@article{mo2026opendecoder,
  title={Opendecoder: Open large language model decoding to incorporate document quality in rag},
  author={Mo, Fengran and Su, Zhan and Hui, Yuchen and Zhang, Jinghan and Sun, Jia Ao and Liu, Zheyuan and Zhang, Chao and Sakai, Tetsuya and Nie, Jian-Yun},
  journal={arXiv preprint arXiv:2601.09028},
  year={2026}
}

@article{huang2026survey,
  title={A survey on large language models with multilingualism: Recent advances and new frontiers},
  author={Huang, Kaiyu and Mo, Fengran and Zhang, Xinyu and Li, Hongliang and Li, You and Zhang, Yuanchi and Yi, Weijian and Mao, Yulong and Liu, Jinchen and Xu, Yuzhuang and others},
  journal={Artificial Intelligence Review},
  year={2026},
  publisher={Springer}
}

@article{mo2026agentic,
  title={Agentic Conversational Search with Contextualized Reasoning via Reinforcement Learning},
  author={Mo, Fengran and Gao, Yifan and Li, Sha and Zeng, Hansi and Liu, Xin and Tan, Zhaoxuan and Li, Xian and Chen, Jianshu and Wang, Dakuo and Jiang, Meng},
  journal={arXiv preprint arXiv:2601.13115},
  year={2026}
}

@article{zeng2026synplanresearch,
  title={SynPlanResearch-R1: Encouraging Tool Exploration for Deep Research with Synthetic Plans},
  author={Zeng, Hansi and Li, Zoey and Gao, Yifan and Zhang, Chenwei and Pan, Xiaoman and Yang, Tao and Mo, Fengran and Lin, Jiacheng and Li, Xian and Shang, Jingbo},
  journal={arXiv preprint arXiv:2603.07853},
  year={2026}
}

@inproceedings{mo2025uniconv,
  title={Uniconv: Unifying retrieval and response generation for large language models in conversations},
  author={Mo, Fengran and Gao, Yifan and Meng, Chuan and Liu, Xin and Wu, Zhuofeng and Mao, Kelong and Wang, Zhengyang and Chen, Pei and Li, Zheng and Li, Xian and others},
  booktitle={Proceedings of the 63rd Annual Meeting of the Association for Computational Linguistics (Volume 1: Long Papers)},
  pages={6936--6949},
  year={2025}
}

@misc{wu2024languagesequalinsightsmultilingual,
      title={Not All Languages are Equal: Insights into Multilingual Retrieval-Augmented Generation}, 
      author={Suhang Wu and Jialong Tang and Baosong Yang and Ante Wang and Kaidi Jia and Jiawei Yu and Junfeng Yao and Jinsong Su},
      year={2024},
      eprint={2410.21970},
      archivePrefix={arXiv},
      primaryClass={cs.CL},
      url={https://arxiv.org/abs/2410.21970}, 
}

@inproceedings{ranaldi-etal-2025-improving-multilingual,
    title = "Improving Multilingual Retrieval-Augmented Language Models through Dialectic Reasoning Argumentations",
    author = "Ranaldi, Leonardo  and
      Ranaldi, Federico  and
      Zanzotto, Fabio Massimo  and
      Haddow, Barry  and
      Birch, Alexandra",
    editor = "Christodoulopoulos, Christos  and
      Chakraborty, Tanmoy  and
      Rose, Carolyn  and
      Peng, Violet",
    booktitle = "Proceedings of the 2025 Conference on Empirical Methods in Natural Language Processing",
    month = nov,
    year = "2025",
    address = "Suzhou, China",
    publisher = "Association for Computational Linguistics",
    url = "https://aclanthology.org/2025.emnlp-main.461/",
    doi = "10.18653/v1/2025.emnlp-main.461",
    pages = "9075--9096",
    ISBN = "979-8-89176-332-6"
}

@misc{ranaldi2025multilingualretrievalaugmentedgenerationknowledgeintensive,
      title={Multilingual Retrieval-Augmented Generation for Knowledge-Intensive Task}, 
      author={Leonardo Ranaldi and Barry Haddow and Alexandra Birch},
      year={2025},
      eprint={2504.03616},
      archivePrefix={arXiv},
      primaryClass={cs.CL},
      url={https://arxiv.org/abs/2504.03616}, 
}

@INPROCEEDINGS{10594504,
  author={Wang, Wenmin and Zhang, Peilin and Liu, Ge and Wu, Ruihua and Song, Guixiang},
  booktitle={2024 IEEE 4th International Conference on Electronic Technology, Communication and Information (ICETCI)}, 
  title={Investigating on the External Knowledge in RAG for Zero-Shot Cross-Language Transfer}, 
  year={2024},
  volume={},
  number={},
  pages={1479-1484},
  doi={10.1109/ICETCI61221.2024.10594504}}

@inproceedings{qi2025sot,
  title={SoT: Structured-of-Thought Prompting Guides Multilingual Reasoning in Large Language Models},
  author={Qi, Rui and Man, Zhibo and Chen, Yufeng and Mo, Fengran and Xu, Jinan and Huang, Kaiyu},
  booktitle={Findings of the Association for Computational Linguistics: EMNLP 2025},
  pages={11024--11039},
  year={2025}
}

@inproceedings{huang-etal-2025-boosting,
    title = "Boosting Data Utilization for Multilingual Dense Retrieval",
    author = "Huang, Chao  and
      Mo, Fengran  and
      Chen, Yufeng  and
      Guan, Changhao  and
      Yue, Zhenrui  and
      Wang, Xinyu  and
      Xu, Jinan  and
      Huang, Kaiyu",
    editor = "Christodoulopoulos, Christos  and
      Chakraborty, Tanmoy  and
      Rose, Carolyn  and
      Peng, Violet",
    booktitle = "Proceedings of the 2025 Conference on Empirical Methods in Natural Language Processing",
    month = nov,
    year = "2025",
    address = "Suzhou, China",
    publisher = "Association for Computational Linguistics",
    url = "https://aclanthology.org/2025.emnlp-main.624/",
    doi = "10.18653/v1/2025.emnlp-main.624",
    pages = "12362--12378",
    ISBN = "979-8-89176-332-6",
}

@article{su2025parametriclora,
  title={Parametric retrieval-augmented generation using latent routing of lora adapters},
  author={Su, Zhan and Mo, Fengran and Zhang, Jinghan and Hui, Yuchen and Sun, Jiaao and Nie, Jian-yun},
  journal={arXiv preprint arXiv:2511.17044},
  year={2025}
}

@inproceedings{su2025parametric,
  title={Parametric retrieval augmented generation},
  author={Su, Weihang and Tang, Yichen and Ai, Qingyao and Yan, Junxi and Wang, Changyue and Wang, Hongning and Ye, Ziyi and Zhou, Yujia and Liu, Yiqun},
  booktitle={Proceedings of the 48th International ACM SIGIR Conference on Research and Development in Information Retrieval},
  pages={1240--1250},
  year={2025}
}

@inproceedings{zhang2024prototypical,
  title={Prototypical reward network for data-efficient rlhf},
  author={Zhang, Jinghan and Wang, Xiting and Jin, Yiqiao and Chen, Changyu and Zhang, Xinhao and Liu, Kunpeng},
  booktitle={Proceedings of the 62nd Annual Meeting of the Association for Computational Linguistics (Volume 1: Long Papers)},
  pages={13871--13884},
  year={2024}
}

@article{mo2025survey,
  title={A survey of conversational search},
  author={Mo, Fengran and Mao, Kelong and Zhao, Ziliang and Qian, Hongjin and Chen, Haonan and Cheng, Yiruo and Li, Xiaoxi and Zhu, Yutao and Dou, Zhicheng and Nie, Jian-Yun},
  journal={ACM Transactions on Information Systems},
  volume={43},
  number={6},
  pages={1--50},
  year={2025},
  publisher={ACM New York, NY}
}

@inproceedings{zhang2025ratt,
  title={Ratt: A thought structure for coherent and correct llm reasoning},
  author={Zhang, Jinghan and Wang, Xiting and Ren, Weijieying and Jiang, Lu and Wang, Dongjie and Liu, Kunpeng},
  booktitle={Proceedings of the AAAI Conference on Artificial Intelligence},
  volume={39},
  number={25},
  pages={26733--26741},
  year={2025}
}

@article{lewis2020retrieval,
  title={Retrieval-augmented generation for knowledge-intensive nlp tasks},
  author={Lewis, Patrick and Perez, Ethan and Piktus, Aleksandra and Petroni, Fabio and Karpukhin, Vladimir and Goyal, Naman and K{\"u}ttler, Heinrich and Lewis, Mike and Yih, Wen-tau and Rockt{\"a}schel, Tim and others},
  journal={Advances in neural information processing systems},
  volume={33},
  pages={9459--9474},
  year={2020}
}

@inproceedings{trivedi2023interleaving,
  title={Interleaving retrieval with chain-of-thought reasoning for knowledge-intensive multi-step questions},
  author={Trivedi, Harsh and Balasubramanian, Niranjan and Khot, Tushar and Sabharwal, Ashish},
  booktitle={Proceedings of the 61st annual meeting of the association for computational linguistics (volume 1: long papers)},
  pages={10014--10037},
  year={2023}
}

@article{li2025search,
  title={Search-o1: Agentic search-enhanced large reasoning models},
  author={Li, Xiaoxi and Dong, Guanting and Jin, Jiajie and Zhang, Yuyao and Zhou, Yujia and Zhu, Yutao and Zhang, Peitian and Dou, Zhicheng},
  journal={arXiv preprint arXiv:2501.05366},
  year={2025}
}

@article{chung2024scaling,
  title={Scaling instruction-finetuned language models},
  author={Chung, Hyung Won and Hou, Le and Longpre, Shayne and Zoph, Barret and Tay, Yi and Fedus, William and Li, Yunxuan and Wang, Xuezhi and Dehghani, Mostafa and Brahma, Siddhartha and others},
  journal={Journal of Machine Learning Research},
  volume={25},
  number={70},
  pages={1--53},
  year={2024}
}

@article{wang2024multilingual,
  title={Multilingual e5 text embeddings: A technical report},
  author={Wang, Liang and Yang, Nan and Huang, Xiaolong and Yang, Linjun and Majumder, Rangan and Wei, Furu},
  journal={arXiv preprint arXiv:2402.05672},
  year={2024}
}

@inproceedings{popqa-etal-2023-trust,
    title = "When Not to Trust Language Models: Investigating Effectiveness of Parametric and Non-Parametric Memories",
    author = "Mallen, Alex  and
      Asai, Akari  and
      Zhong, Victor  and
      Das, Rajarshi  and
      Khashabi, Daniel  and
      Hajishirzi, Hannaneh",
    editor = "Rogers, Anna  and
      Boyd-Graber, Jordan  and
      Okazaki, Naoaki",
    booktitle = "Proceedings of the 61st Annual Meeting of the Association for Computational Linguistics (Volume 1: Long Papers)",
    month = jul,
    year = "2023",
    address = "Toronto, Canada",
    publisher = "Association for Computational Linguistics",
    url = "https://aclanthology.org/2023.acl-long.546/",
    doi = "10.18653/v1/2023.acl-long.546",
    pages = "9802--9822"
}

@inproceedings{yang-etal-2018-hotpotqa,
    title = "{H}otpot{QA}: A Dataset for Diverse, Explainable Multi-hop Question Answering",
    author = "Yang, Zhilin  and
      Qi, Peng  and
      Zhang, Saizheng  and
      Bengio, Yoshua  and
      Cohen, William  and
      Salakhutdinov, Ruslan  and
      Manning, Christopher D.",
    editor = "Riloff, Ellen  and
      Chiang, David  and
      Hockenmaier, Julia  and
      Tsujii, Jun{'}ichi",
    booktitle = "Proceedings of the 2018 Conference on Empirical Methods in Natural Language Processing",
    month = oct # "-" # nov,
    year = "2018",
    address = "Brussels, Belgium",
    publisher = "Association for Computational Linguistics",
    url = "https://aclanthology.org/D18-1259/",
    doi = "10.18653/v1/D18-1259",
    pages = "2369--2380"
}

@inproceedings{ho-etal-2020-2wiki,
    title = "Constructing A Multi-hop {QA} Dataset for Comprehensive Evaluation of Reasoning Steps",
    author = "Ho, Xanh  and
      Duong Nguyen, Anh-Khoa  and
      Sugawara, Saku  and
      Aizawa, Akiko",
    editor = "Scott, Donia  and
      Bel, Nuria  and
      Zong, Chengqing",
    booktitle = "Proceedings of the 28th International Conference on Computational Linguistics",
    month = dec,
    year = "2020",
    address = "Barcelona, Spain (Online)",
    publisher = "International Committee on Computational Linguistics",
    url = "https://aclanthology.org/2020.coling-main.580/",
    doi = "10.18653/v1/2020.coling-main.580",
    pages = "6609--6625"
}

@misc{qwen2025qwen25technicalreport,
      title={Qwen2.5 Technical Report}, 
      author={Qwen and : and An Yang and Baosong Yang and Beichen Zhang and Binyuan Hui and Bo Zheng and Bowen Yu and Chengyuan Li and Dayiheng Liu and Fei Huang and Haoran Wei and Huan Lin and Jian Yang and Jianhong Tu and Jianwei Zhang and Jianxin Yang and Jiaxi Yang and Jingren Zhou and Junyang Lin and Kai Dang and Keming Lu and Keqin Bao and Kexin Yang and Le Yu and Mei Li and Mingfeng Xue and Pei Zhang and Qin Zhu and Rui Men and Runji Lin and Tianhao Li and Tianyi Tang and Tingyu Xia and Xingzhang Ren and Xuancheng Ren and Yang Fan and Yang Su and Yichang Zhang and Yu Wan and Yuqiong Liu and Zeyu Cui and Zhenru Zhang and Zihan Qiu},
      year={2025},
      eprint={2412.15115},
      archivePrefix={arXiv},
      primaryClass={cs.CL},
      url={https://arxiv.org/abs/2412.15115}, 
}

@inproceedings{liu-etal-2025-xrag,
    title = "{XRAG}: Cross-lingual Retrieval-Augmented Generation",
    author = "Liu, Wei  and
      Trenous, Sony  and
      Ribeiro, Leonardo F. R.  and
      Byrne, Bill  and
      Hieber, Felix",
    editor = "Christodoulopoulos, Christos  and
      Chakraborty, Tanmoy  and
      Rose, Carolyn  and
      Peng, Violet",
    booktitle = "Findings of the Association for Computational Linguistics: EMNLP 2025",
    month = nov,
    year = "2025",
    address = "Suzhou, China",
    publisher = "Association for Computational Linguistics",
    url = "https://aclanthology.org/2025.findings-emnlp.849/",
    doi = "10.18653/v1/2025.findings-emnlp.849",
    pages = "15669--15690",
    ISBN = "979-8-89176-335-7"
}

@inproceedings{10.1145/3726302.3730201,
author = {Lawrie, Dawn and Kayi, Efsun and Yang, Eugene and Mayfield, James and Oard, Douglas W. and Miller, Scott},
title = {Generate-Distill: Training Cross-Language IR Models with Synthetically-Generated Data},
year = {2025},
isbn = {9798400715921},
publisher = {Association for Computing Machinery},
address = {New York, NY, USA},
url = {https://doi.org/10.1145/3726302.3730201},
doi = {10.1145/3726302.3730201},
booktitle = {Proceedings of the 48th International ACM SIGIR Conference on Research and Development in Information Retrieval},
pages = {2926–2930},
numpages = {5},
location = {Padua, Italy},
series = {SIGIR '25}
}

@misc{shao2024deepseekmathpushinglimitsmathematical,
      title={DeepSeekMath: Pushing the Limits of Mathematical Reasoning in Open Language Models}, 
      author={Zhihong Shao and Peiyi Wang and Qihao Zhu and Runxin Xu and Junxiao Song and Xiao Bi and Haowei Zhang and Mingchuan Zhang and Y. K. Li and Y. Wu and Daya Guo},
      year={2024},
      eprint={2402.03300},
      archivePrefix={arXiv},
      primaryClass={cs.CL},
      url={https://arxiv.org/abs/2402.03300}, 
}

@misc{hossain2026costefficientcrosslingualretrievalaugmentedgeneration,
      title={Cost-Efficient Cross-Lingual Retrieval-Augmented Generation for Low-Resource Languages: A Case Study in Bengali Agricultural Advisory}, 
      author={Md. Asif Hossain and Nabil Subhan and Mantasha Rahman Mahi and Jannatul Ferdous Nabila},
      year={2026},
      eprint={2601.02065},
      archivePrefix={arXiv},
      primaryClass={cs.CL},
      url={https://arxiv.org/abs/2601.02065}, 
}

@inproceedings{10.1145/3726302.3729921,
author = {Cui, Chenxu and Fan, Haihui and Zhang, Jinchao and Shen, Lin and Li, Bo and Wang, Weiping},
title = {CIRAG: Retrieval-Augmented Language Model with Collective Intelligence},
year = {2025},
isbn = {9798400715921},
publisher = {Association for Computing Machinery},
address = {New York, NY, USA},
url = {https://doi.org/10.1145/3726302.3729921},
doi = {10.1145/3726302.3729921},
booktitle = {Proceedings of the 48th International ACM SIGIR Conference on Research and Development in Information Retrieval},
pages = {1316–1326},
numpages = {11},
location = {Padua, Italy},
series = {SIGIR '25}
}

@inproceedings{10.1145/3726302.3729920,
author = {Hu, Yuntong and Lei, Zhihan and Dai, Zhongjie and Zhang, Allen and Angirekula, Abhinav and Zhang, Zheng and Zhao, Liang},
title = {CG-RAG: Research Question Answering by Citation Graph Retrieval-Augmented LLMs},
year = {2025},
isbn = {9798400715921},
publisher = {Association for Computing Machinery},
address = {New York, NY, USA},
url = {https://doi.org/10.1145/3726302.3729920},
doi = {10.1145/3726302.3729920},
booktitle = {Proceedings of the 48th International ACM SIGIR Conference on Research and Development in Information Retrieval},
pages = {678–687},
numpages = {10},
location = {Padua, Italy},
series = {SIGIR '25}
}

@inproceedings{10.1145/3726302.3729907,
author = {Tang, Yubao and Zhang, Ruqing and Guo, Jiafeng and de Rijke, Maarten and Fan, Yixing and Cheng, Xueqi},
title = {Boosting Retrieval-Augmented Generation with Generation-Augmented Retrieval: A Co-Training Approach},
year = {2025},
isbn = {9798400715921},
publisher = {Association for Computing Machinery},
address = {New York, NY, USA},
url = {https://doi.org/10.1145/3726302.3729907},
doi = {10.1145/3726302.3729907},
booktitle = {Proceedings of the 48th International ACM SIGIR Conference on Research and Development in Information Retrieval},
pages = {2441–2451},
numpages = {11},
location = {Padua, Italy},
series = {SIGIR '25}
}

@inproceedings{10.1145/3626772.3657819,
author = {Guo, Ping and Ren, Yubing and Hu, Yue and Cao, Yanan and Li, Yunpeng and Huang, Heyan},
title = {Steering Large Language Models for Cross-lingual Information Retrieval},
year = {2024},
isbn = {9798400704314},
publisher = {Association for Computing Machinery},
address = {New York, NY, USA},
url = {https://doi.org/10.1145/3626772.3657819},
doi = {10.1145/3626772.3657819},
booktitle = {Proceedings of the 47th International ACM SIGIR Conference on Research and Development in Information Retrieval},
pages = {585–596},
numpages = {12},
location = {Washington DC, USA},
series = {SIGIR '24}
}

@inproceedings{10.1145/3626772.3657867,
author = {Iana, Andreea and Glava\v{s}, Goran and Paulheim, Heiko},
title = {MIND Your Language: A Multilingual Dataset for Cross-lingual News Recommendation},
year = {2024},
isbn = {9798400704314},
publisher = {Association for Computing Machinery},
address = {New York, NY, USA},
url = {https://doi.org/10.1145/3626772.3657867},
doi = {10.1145/3626772.3657867},
booktitle = {Proceedings of the 47th International ACM SIGIR Conference on Research and Development in Information Retrieval},
pages = {553–563},
numpages = {11},
location = {Washington DC, USA},
series = {SIGIR '24}
}

@inproceedings{10.1145/3626772.3657760,
author = {Yang, Diji and Rao, Jinmeng and Chen, Kezhen and Guo, Xiaoyuan and Zhang, Yawen and Yang, Jie and Zhang, Yi},
title = {IM-RAG: Multi-Round Retrieval-Augmented Generation Through Learning Inner Monologues},
year = {2024},
isbn = {9798400704314},
publisher = {Association for Computing Machinery},
address = {New York, NY, USA},
url = {https://doi.org/10.1145/3626772.3657760},
doi = {10.1145/3626772.3657760},
booktitle = {Proceedings of the 47th International ACM SIGIR Conference on Research and Development in Information Retrieval},
pages = {730–740},
numpages = {11},
location = {Washington DC, USA},
series = {SIGIR '24}
}

@misc{nllbteam2022languageleftbehindscaling,
      title={No Language Left Behind: Scaling Human-Centered Machine Translation}, 
      author={NLLB Team and Marta R. Costa-jussà and James Cross and Onur Çelebi and Maha Elbayad and Kenneth Heafield and Kevin Heffernan and Elahe Kalbassi and Janice Lam and Daniel Licht and Jean Maillard and Anna Sun and Skyler Wang and Guillaume Wenzek and Al Youngblood and Bapi Akula and Loic Barrault and Gabriel Mejia Gonzalez and Prangthip Hansanti and John Hoffman and Semarley Jarrett and Kaushik Ram Sadagopan and Dirk Rowe and Shannon Spruit and Chau Tran and Pierre Andrews and Necip Fazil Ayan and Shruti Bhosale and Sergey Edunov and Angela Fan and Cynthia Gao and Vedanuj Goswami and Francisco Guzmán and Philipp Koehn and Alexandre Mourachko and Christophe Ropers and Safiyyah Saleem and Holger Schwenk and Jeff Wang},
      year={2022},
      eprint={2207.04672},
      archivePrefix={arXiv},
      primaryClass={cs.CL},
      url={https://arxiv.org/abs/2207.04672}, 
}

@INPROCEEDINGS{10756977,
  author={Iscan, Can and Ozara, Muhammet Furkan and Akbulut, Akhan},
  booktitle={2024 Innovations in Intelligent Systems and Applications Conference (ASYU)}, 
  title={Enhancing RAG Pipeline Performance with Translation-Based Embedding Strategies for Non-English Documents}, 
  year={2024},
  volume={},
  number={},
  pages={1-6},
  doi={10.1109/ASYU62119.2024.10756977}}

@misc{ahmad2024enhancingmultilingualinformationretrieval,
      title={Enhancing Multilingual Information Retrieval in Mixed Human Resources Environments: A RAG Model Implementation for Multicultural Enterprise}, 
      author={Syed Rameel Ahmad},
      year={2024},
      eprint={2401.01511},
      archivePrefix={arXiv},
      primaryClass={cs.IR},
      url={https://arxiv.org/abs/2401.01511}, 
}

@misc{hossain2026costefficient,
      title={Cost-Efficient Cross-Lingual Retrieval-Augmented Generation for Low-Resource Languages: A Case Study in Bengali Agricultural Advisory}, 
      author={Md. Asif Hossain and Nabil Subhan and Mantasha Rahman Mahi and Jannatul Ferdous Nabila},
      year={2026},
      eprint={2601.02065},
      archivePrefix={arXiv},
      primaryClass={cs.CL},
      url={https://arxiv.org/abs/2601.02065}, 
}

@misc{li2025languagedriftmultilingualretrievalaugmented,
      title={Language Drift in Multilingual Retrieval-Augmented Generation: Characterization and Decoding-Time Mitigation}, 
      author={Bo Li and Zhenghua Xu and Rui Xie},
      year={2025},
      eprint={2511.09984},
      archivePrefix={arXiv},
      primaryClass={cs.CL},
      url={https://arxiv.org/abs/2511.09984}, 
}


\end{document}